\documentclass[openacc]{rstransa}

\usepackage[nopar]{kantlipsum}
\usepackage[plain]{algorithm}
\usepackage{algpseudocode}

\usepackage{pgfplots}
\DeclareUnicodeCharacter{2212}{−}
\usepgfplotslibrary{groupplots,dateplot}
\pgfplotsset{compat=newest}
\usepackage{tikz}
\usetikzlibrary{positioning, backgrounds, decorations, arrows, arrows.meta, patterns, shapes, fit}
\tikzset{>=latex}
\usepackage{forest}

\usepackage{subcaption}
\usepackage{tabularray}
\usepackage{siunitx}
\titlehead{Research}

\begin{document}

\title{Zobrist Hash-based Duplicate Detection in Symbolic Regression}
\author{Bogdan Burlacu}

\address{University of Applied Sciences Upper Austria\\
Heuristic and Evolutionary Algorithms Laboratory\\
Softwarepark 11, 4232, Hagenberg, Austria}

\subject{machine learning, symbolic regression}


\corres{Bogdan Burlacu\\
\email{bogdan.burlacu@fh-ooe.at}}

\def\CPP{C\raisebox{0.5ex}{\tiny\textbf{++}}}

\begin{abstract}
    Symbolic regression encompasses a family of search algorithms that aim to discover the best fitting function for a set of data without requiring an \emph{a priori} specification of the model structure. The most successful and commonly used technique for symbolic regression is Genetic Programming (GP), an evolutionary search method that evolves a population of mathematical expressions through the mechanism of natural selection. In this work we analyze the efficiency of the evolutionary search in GP and show that many points in the search space are re-visited and re-evaluated multiple times by the algorithm, leading to wasted computational effort. We address this issue by introducing a caching mechanism based on the Zobrist hash, a type of hashing frequently used in abstract board games for the efficient construction and subsequent update of transposition tables. We implement our caching approach using the open-source framework Operon and demonstrate its performance on a selection of real-world regression problems, where we observe up to 34\% speedups without any detrimental effects on search quality. The hashing approach represents a straightforward way to improve runtime performance while also offering some interesting possibilities for adjusting search strategy based on cached information.
\end{abstract}





\maketitle

\section{Introduction}

Symbolic Regression (SR) has been rapidly emerging as a key approach in machine learning in recent years due to the widespread acceptance of interpretability as an unequivocally crucial element of trust in AI~\cite{rudin2022}. Compared to ``black-box'' models such as deep neural networks or support vector machines with nonlinear kernels, whose decision-making processes are opaque and not easily understood by humans, models produced by SR are given in the form of mathematical expressions that are, to a large extent, fathomable, simple and insightful.
Its ``white-box'' characteristics make SR an invaluable tool for scientific discovery in many application domains, such as astrophysics~\cite{matchev2022,bartlett2024}, physical law discovery~\cite{desmond2023,lemos2023,grayeli2024}, materials science~\cite{wang2019,wang2022,wang2024}, particle kinematics~\cite{dong2023,morales-alvarado2024}, robotics~\cite{zhang2023}, controller design~\cite{danai2021}, or clinical decision support~\cite{lacava2023}.
SR distinguishes itself from other regression methods that are also capable of producing symbolic models (e.g. polynomial models up to a given degree) by its ability to learn from data without requiring an \emph{a priori} specification of the model structure. In SR, both model structure and parameters are simultaneously developed, allowing for much greater expressive power and flexibility.

Despite a recent emergence of many competing approaches, based on deep learning~\cite{petersen2021,kamienny2022,landajuela2022,vastl2024}, neural networks~\cite{sahoo2018}, grammar enumeration~\cite{worm2013}, mixed-integer nonlinear programming~\cite{cozad2018}, path-wise regularized learning\cite{mcconaghy2011}, or exhaustive search~\cite{kammerer2020,bartlett2024}, Genetic Programming~\cite{koza1992} (GP), a biologically-inspired metaheuristic operating after the model of natural selection, remains the most successful and overall best performing approach for SR~\cite{lacava2021,defranca2024}.
However, while undeniably powerful, GP-based SR still suffers from the intrinsic limitations of evolutionary methods, such as high computational cost, search inefficiency~\cite{kronberger2024}, diversity loss and risk of premature convergence~\cite{burks2015,chen2021}. These limitations are further compounded by the non-deterministic nature of the search which makes it necessary to perform multiple algorithmic runs to get reliable estimates.

In this work we introduce an approach for mitigating these limitations by caching duplicate expressions encountered during the search, precluding the need for solution re-evaluation and saving execution time. Our method is based on the Zobrist hash, commonly used to detect transpositions in board games like Chess and Go. In our case, a transposition is defined as a sequence of recombination operations (crossover or mutation) that leads to an already-seen expression. In what follows, we show that transpositions occur frequently in SR.

The paper is structured as follows. Section~\ref{sec:genetic-programming} offers a more detailed discussion of GP and argues why duplicates do occur during the search. Section~\ref{sec:duplicate-detection} explains the proposed duplicate detection mechanism and its implementation. Section~\ref{sec:results} provides empirical evidence in favor of fitness caching as a way to improve runtime, also noting a change in search dynamics with caching. Finally, Section~\ref{sec:conclusion} discusses the implications of this work and future research directions.

\section{Symbolic Regression with Genetic Programming}\label{sec:genetic-programming}

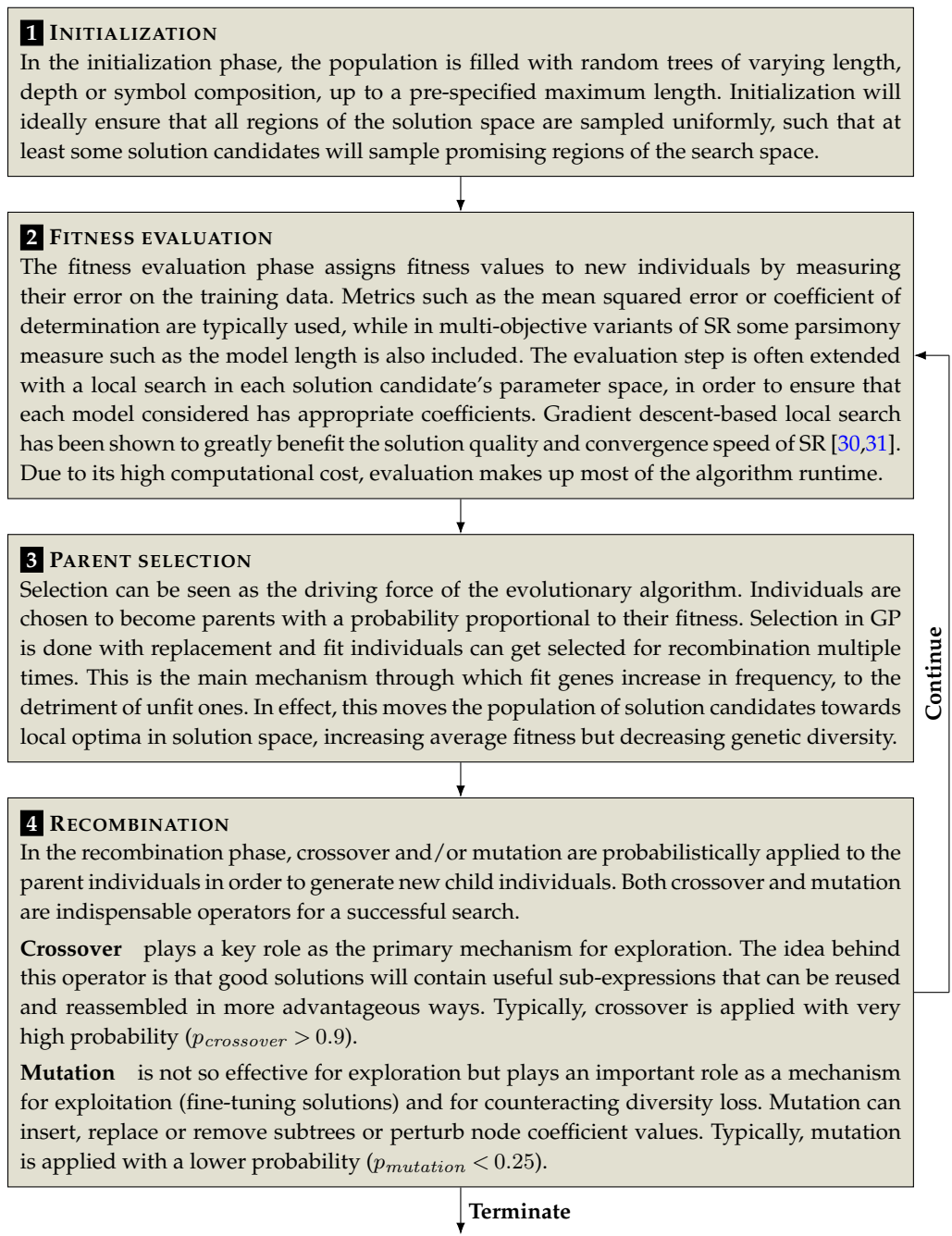
\begin{figure}
    \begin{tikzpicture}[node distance=0.5cm and 0.5cm]
    \newcommand*\circled[1]{\tikz[baseline=(char.base)]{
            \node[shape=rectangle,draw,fill=black,text=white,inner sep=2pt] (char) {#1};}}
    \tikzset{mynode/.style={align=justify, text width=0.928\textwidth, inner sep=5pt, draw, fill=abstractcolor}}
    \node[mynode] (a) { \textsc{\textbf{\circled{1} Initialization}}\\
        In the initialization phase, the population is filled with random trees of varying length, depth or symbol composition, up to a pre-specified maximum length. Initialization will ideally ensure that all regions of the solution space are sampled uniformly, such that at least some solution candidates will sample promising regions of the search space.
    };

    \node[mynode, below=of a] (b) { \textsc{\textbf{\circled{2} Fitness evaluation}}\\
        The fitness evaluation phase assigns fitness values to new individuals by measuring their error on the training data. Metrics such as the mean squared error or coefficient of determination are typically used, while in multi-objective variants of SR some parsimony measure such as the model length is also included. The evaluation step is often extended with a local search in each solution candidate's parameter space, in order to ensure that each model considered has appropriate coefficients. Gradient descent-based local search has been shown to greatly benefit the solution quality and convergence speed of SR~\cite{kommenda2020,burlacu2024}. Due to its high computational cost, evaluation makes up most of the algorithm runtime.
    };

    \node[mynode, below=of b] (c) { \textsc{\textbf{\circled{3} Parent selection}}\\
        Selection can be seen as the driving force of the evolutionary algorithm. Individuals are chosen to become parents with a probability proportional to their fitness. Selection in GP is done with replacement and fit individuals can get selected for recombination multiple times. This is the main mechanism through which fit genes increase in frequency, to the detriment of unfit ones. In effect, this moves the population of solution candidates towards local optima in solution space, increasing average fitness but decreasing genetic diversity.
    };

    \node[mynode, below=of c] (d) { \textsc{\textbf{\circled{4} Recombination}}\\
        In the recombination phase, crossover and/or mutation are probabilistically applied to the parent individuals in order to generate new child individuals. Both crossover and mutation are indispensable operators for a successful search.\vspace{-1.2em}
        \paragraph{Crossover} plays a key role as the primary mechanism for exploration. The idea behind this operator is that good solutions will contain useful sub-expressions that can be reused and reassembled in more advantageous ways. Typically, crossover is applied with very high probability ($p_\mathit{crossover} > 0.9$).\vspace{-1.2em}
        \paragraph{Mutation} is not so effective for exploration but plays an important role as a mechanism for exploitation (fine-tuning solutions) and for counteracting diversity loss. Mutation can insert, replace or remove subtrees or perturb node coefficient values. Typically, mutation is applied with a lower probability ($p_\mathit{mutation} < 0.25$).
    };

    \node[below=of d, yshift=-5pt, minimum height=0pt, minimum width=0pt, text width=0pt] (e) {};

    \node[right=of b, inner sep=0pt, outer sep=0pt, minimum size=0pt] (x1) {};
    \node[right=of d, inner sep=0pt, outer sep=0pt, minimum size=0pt] (x2) {};


    \draw[-Latex] (a) -- (b);
    \draw[-Latex] (b) -- (c);
    \draw[-Latex] (c) -- (d);
    \draw[-Latex] (d) -- node[minimum height=0pt, minimum width=0pt, text width=46pt, anchor=west]{ \textbf{Terminate} } (e);
    \draw (d.east) -- ++(0.5cm,0) (x1.center);
    \draw[Latex-] (b.east) -- ++(0.5cm,0) (x2);
    \draw (x1) -- (x2) node[label={[rotate=90, midway]\textbf{Continue}}] {} ;
\end{tikzpicture}\vspace{-2em}
    \caption{Evolutionary workflow of Genetic Programming}
    \label{fig:gp-workflow}
\end{figure}

Genetic Programming~\cite{koza1992} is a metaheuristic optimization method that performs a \emph{global search} in solution space by orchestrating the evolution of a \textit{population} of computer programs using the basic mechanisms of \textit{selection} and \textit{recombination}.
Each program, called an \textit{individual}, in keeping with the biological metaphor, contains instructions for solving the problem at hand and its \textit{fitness} measures its ability to achieve the problem objectives\footnote{The actual representation and fitness measure are problem-dependent.}. Starting with a random initial population, in each iteration of the algorithm, the fittest individuals are selected for reproduction and subsequently swap parts of their structure to produce potentially fitter offspring. The process continues for a fixed number of iterations or until convergence, according to the workflow described in~\autoref{fig:gp-workflow}.

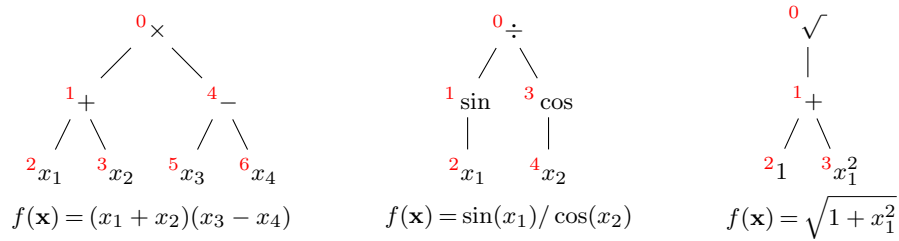
\begin{figure}
    \centering
    \begin{tikzpicture}
    \node[align=center] (n1) {
        \begin{forest}
            [$^{\color{red} 0}\times$ [$^{\color{red} 1}+$ [$^{\color{red}2}x_1$] [$^{\color{red}3}x_2$]] [$^{\color{red}4}-$ [$^{\color{red}5}x_3$] [$^{\color{red}6}x_4$]]]
        \end{forest}\\

        $f(\mathbf{x}) = (x_1 + x_2)(x_3 - x_4)$
    };

    \node[align=center, right=of n1] (n2) {
        \begin{forest}
            [ $^{\color{red} 0}\div$
                [ $^{\color{red} 1}\sin$ [$^{\color{red} 2}x_1$] ]
                [ $^{\color{red} 3}\cos$ [$^{\color{red} 4}x_2$] ]
            ]
        \end{forest}\\

        $f(\mathbf{x}) = \sin(x_1)/\cos(x_2)$
    };

    \node[align=center, right=of n2] (n3) {
        \begin{forest}
            [ $^{\color{red} 0}\sqrt{~}$
                [ $^{\color{red} 1}+$
                    [ $^{\color{red} 2}1$ ]
                    [ $^{\color{red} 3}x_1^2$ ]
                ]
            ]
        \end{forest}\\
        $f(\mathbf{x}) = \sqrt{1 + x_1^2}$
    };
\end{tikzpicture}
    \caption{Example tree representations with preorder traversal indices shown in red.}
    \label{fig:sr-tree-example}
\end{figure}

In GP-based SR, the tree representation encodes mathematical expressions, as illustrated in~\autoref{fig:sr-tree-example}. SR performs a search of the space of mathematical expressions by gradually assembling tree structures from a pre-configured set of functions and terminals.
Given a set of binary (e.g. $+, -, \times, \div, \mathrm{pow}, \mathrm{aq}, \dots$), unary (e.g. $\sin, \cos, \exp, \log, \sqrt{\cdot}, \dots$) and nullary (i.e. variables and parameters) operators, the search space is defined as all valid compositions of operators (subject to some size constraints), making it a discrete combinatorial search space whose dimension scales exponentially with the number of operators.

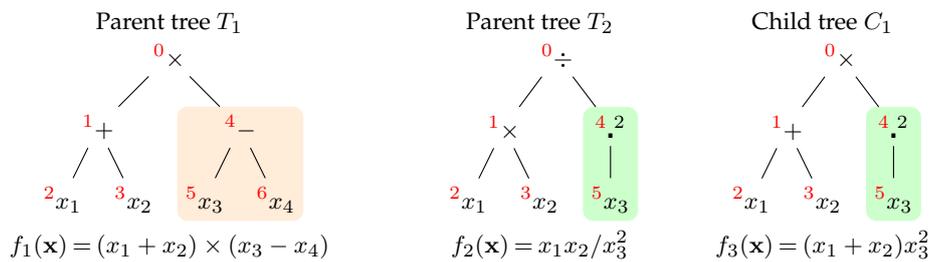
\begin{figure}
    \begin{subfigure}[b]{0.42\textwidth}
    \centering
    Parent tree $T_1$\\
    \begin{forest}
        [{$^{\color{red} 0}\times$}, name=n0
            [{$^{\color{red} 1}+$}, name=n1
                [{$^{\color{red}2}x_1$}, name=n2]
                [{$^{\color{red}3}x_2$}, name=n3]
            ]
            [{$^{\color{red}4}-$}, name=n4
                [{$^{\color{red}5}x_3$}, name=n5]
                [{$^{\color{red}6}x_4$}, name=n6]
            ]
        ]
        \begin{scope}[on background layer]
            \node[fill=orange!15, inner sep=0pt, rectangle, rounded corners, fit=(n4) (n5) (n6)] {};
        \end{scope}
    \end{forest}\\
    $f_1(\mathbf{x}) = (x_1 + x_2) \times (x_3 - x_4)$
\end{subfigure}
\begin{subfigure}[b]{0.29\textwidth}
    \centering
    Parent tree $T_2$\\
    \begin{forest}
        [{$^{\color{red} 0}\div$}, name=n0
            [{$^{\color{red} 1}\times$}, name=n1
                [{$^{\color{red}2}x_1$}, name=n2]
                [{$^{\color{red}3}x_2$}, name=n3]
            ]
            [{$^{\color{red}4}\centerdot^2$}, name=n4
                [{$^{\color{red}5}x_3$}, name=n5]
            ]
        ]
        \begin{scope}[on background layer]
            \node[fill=green!20, inner sep=0pt, rectangle, rounded corners, fit=(n4) (n5)] {};
        \end{scope}
    \end{forest}\\
    $f_2(\mathbf{x}) = {x_1 x_2}/{x_3^2}$
\end{subfigure}
\begin{subfigure}[b]{0.25\textwidth}
    \centering
    Child tree $C_1$\\
    \begin{forest}
        [{$^{\color{red} 0}\times$}, name=n0
            [{$^{\color{red} 1}+$}, name=n1
                [{$^{\color{red}2}x_1$}, name=n2]
                [{$^{\color{red}3}x_2$}, name=n3]
            ]
            [{$^{\color{red}4}\centerdot^2$}, name=n4
                [{$^{\color{red}5}x_3$}, name=n5]
            ]
        ]
        \begin{scope}[on background layer]
            \node[fill=green!20, inner sep=0pt, rectangle, rounded corners, fit=(n4) (n5)] {};
        \end{scope}
    \end{forest}\\
    $f_3(\mathbf{x}) = (x_1 + x_2){x_3^2}$
\end{subfigure}
    \caption{Crossover between two parent tree-encoded expressions. A new symbolic expression $f_3(\mathbf{x}) = (x_1 + x_2){x_3^2}$ is obtained by exchanging the parts $(x_3-x_4)$ and $x_3^2$ between the parents.}\label{fig:gp-crossover}
\end{figure}

The successful exploration of the search space depends on the existence of a sufficiently diverse set of individuals in the population, from which adaptive change can emerge as the product of recombination. At the same time, successful \textit{exploitation}, understood as the local improvement of solutions in the vicinity of optima points, requires a focusing of the search effort and depends on the selection mechanism's ability to concentrate the population on promising regions of the search space.
A balance between these two antagonistic aspects of the search is difficult to achieve since it requires a deep understanding of evolutionary dynamics and is, in general, problem-dependent~\cite{crepinsek2013}. A good trade-off depends on appropriate settings for parameters such as the population size, crossover and mutation probabilities or selection pressure. The latter plays a particularly important role, since it controls the survival and reproduction rates of fit individuals.

\begin{figure}
    \centering
    \includegraphics[width=0.8\textwidth]{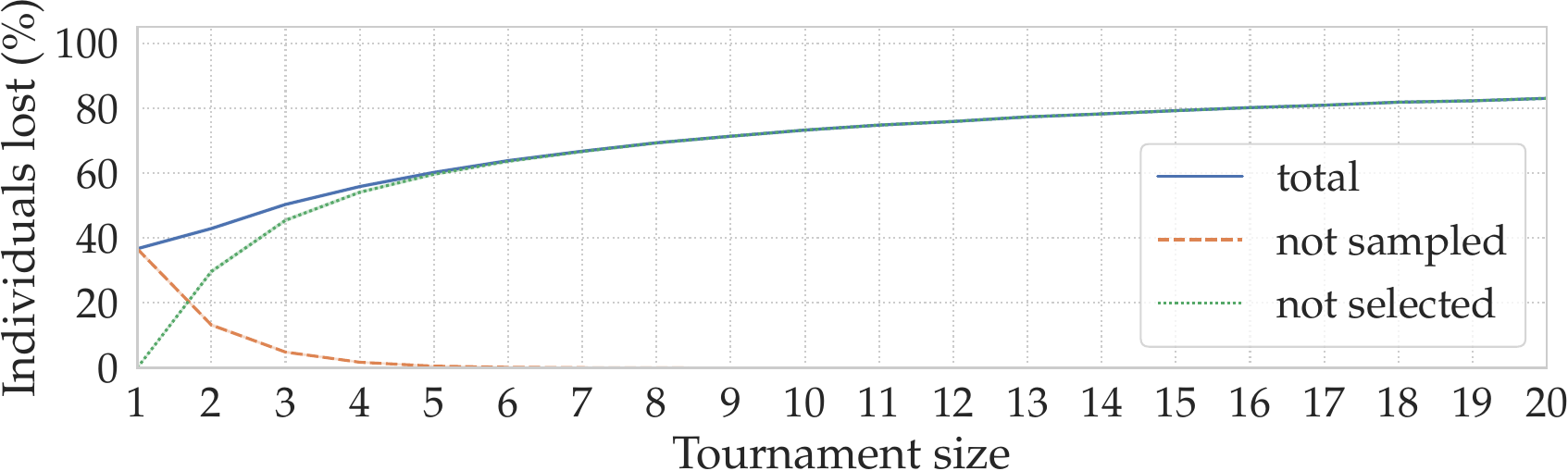}
    \caption{Percentage of not selected individuals as a function of tournament size}\label{fig:tournament-selection-pressure}
\end{figure}

Selection in a finite population causes diversity loss due to duplication of fit individuals\cite{prugel-bennett2000}. In tournament selection, the most popular selection mechanism in SR, this effect was shown to be significant even at small tournament sizes\cite{xie2011}. A simulation of tournament selection in \autoref{fig:tournament-selection-pressure} shows that at least 40\% of individuals are lost in every generation.
Diversity has an impact on search convergence, as the outcome of recombination will depend on the amount of genetic material shared by the parents. The more similar the parents, the more likely a significant amount of duplicate material will be inherited by the child.

From a practical perspective, given the unavoidable loss of diversity over the course of evolution, it is reasonable to assume that some solutions produced by recombination will not be unique, and their initial evaluations could be reused, with significant runtime benefits. Furthermore, detecting when duplicates \textit{may} occur or detecting the \textit{frequency} of duplicate individuals in the population may also lead to better strategies for diversity maintenance. For these reasons, a practical solution for \emph{memoization} in SR can bring significant benefits as it would be able to exploit the inherent duplication and waste of computational effort within GP.

Previous attempts to improve runtime efficiency focused on alternative representations, for example storing the population as a graph\cite{handley1994,keijzer2004,defranca2025} or subtree caching\cite{wong2008}. The caching approach is of particular interest here as it does not require changes to the fundamental structure of the algorithm. However, implementing an efficient cache requires an accurate, low-overhead hashing function, with certain desirable properties such as hash involution. This paper introduces a hashing approach based on Zobrist keys that allows the caching of expression trees with minimal overhead, whose inner workings are described in the following section.

\section{Duplicate Detection in Symbolic Regression}\label{sec:duplicate-detection}

Duplicate detection in SR populations is challenging because of the need to compute distances between unordered labeled trees, a problem known to be NP-complete\cite{zhang1992}. However, if one relaxes the problem to approximate matching, then hash-based tree isomorphism offers similar results with a marginal error bounded by the collision rate\footnote{Perfect hashes are not considered due to their high computational cost.}. Hash functions based on modular arithmetic\cite{wong2008} or Merkle trees\cite{burlacu2019} have been previously used in SR to improve runtime and detect structural and semantic duplicates, but due to their logarithmic time complexity, their overhead remains significant.

Zobrist hashing, named after its inventor Albert Zobrist, is a hashing technique primarily used in computer game algorithms, particularly for board games such as Chess and Go.
At its core, Zobrist hashing involves assigning a unique random number to each possible piece on each possible square of the game board. For instance, in chess, each combination of piece type and board position is mapped to a distinct random number. The hash value of a particular board state is then computed by performing a bitwise XOR (exclusive OR) operation on the numbers corresponding to the pieces present on the board.

The Zobrist hash has excellent runtime properties as XOR operations are generally very well-optimized on modern CPUs due to their simplicity and suitability for parallel execution. But the main advantage of Zobrist hashing becomes apparent when considering the mathematical properties of the XOR function, namely associativity, commutativity and involution.
\begin{align*}
    a \oplus b &= b \oplus a & & \text{(associativity)}\\
    a \oplus (b \oplus c) &= (a \oplus b) \oplus c & & \text{(commutativity)}\\
    a \oplus b \oplus b &= a & & \text{(involution)}\\
\end{align*}
Given an abstract board game whose state can be reached through different move orders, the XOR function enables the incremental update of the game state by only considering changes to the board position after a move is made. In other words, previously visited states can be identified regardless of move order. This helps tree search algorithms such as \emph{Minimax} avoid already-visited nodes in the search tree, by maintaining a data structure called a \emph{transposition table} -- a database that stores results of previously performed searches -- greatly improving search efficiency.

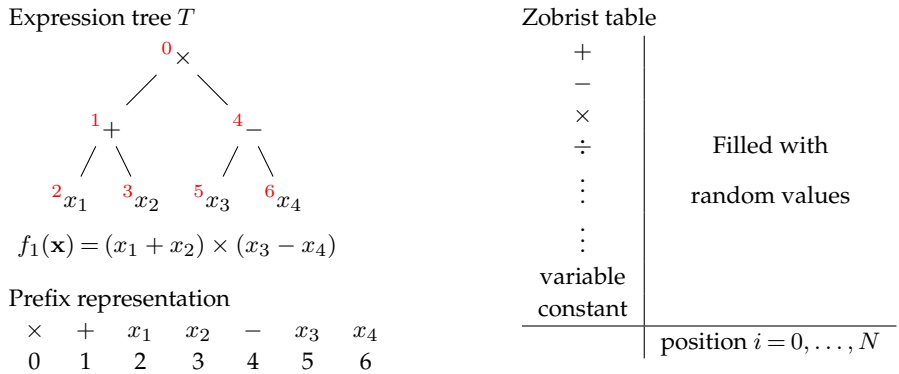
\begin{figure}
    \begin{subfigure}[b][3cm]{0.5\textwidth}
    Expression tree $T$

    \begin{forest}
        [{$^{\color{red} 0}\times$}, name=n0
            [{$^{\color{red} 1}+$}, name=n1
                [{$^{\color{red}2}x_1$}, name=n2]
                [{$^{\color{red}3}x_2$}, name=n3]
            ]
            [{$^{\color{red}4}-$}, name=n4
                [{$^{\color{red}5}x_3$}, name=n5]
                [{$^{\color{red}6}x_4$}, name=n6]
            ]
        ]
        \node[below=of n0, yshift=-1cm] {
            $f_1(\mathbf{x}) = (x_1 + x_2) \times (x_3 - x_4)$
        };
    \end{forest}
    \smallskip

    Prefix representation

    \begin{tabular}{ccccccc}
        $\times$ & $+$ & $x_1$ & $x_2$ & $-$ & $x_3$ & $x_4$\\
        0 & 1 & 2 & 3 & 4 & 5 & 6\\
    \end{tabular}

    \vspace{-2.33cm}
\end{subfigure}\begin{subfigure}[b][3cm]{0.5\textwidth}
    Zobrist table\smallskip

    \begin{tabular}{c|c}
        $+$ & \\
        $-$ & \\
        $\times$ & \\
        $\div$ & Filled with\\
        $\vdots$ & random values\\
        $\vdots$ & \\
        $\text{variable}$ & \\
        $\text{constant}$ & \\
        \hline
        & position $i = 0,\dots,N$\\
    \end{tabular}
\end{subfigure}\vspace{1.5em}
    \caption{$\mathcal{H}(T) = h_{\times,0} \oplus h_{+, 1} \oplus h_{x_1, 2} \oplus h_{x_2, 3} \oplus h_{-,4} \oplus h_{x_3, 5} \oplus h_{x_4, 6}$}
    \label{fig:zobrist-tree-example}
\end{figure}

The same approach is applicable to SR, considering the tree node type and its preorder index as the tabular coordinates for retrieving the associated Zobrist key, as illustrated in \autoref{fig:zobrist-tree-example}. When two parent trees are crossed-over, the child hash will not be recomputed but instead updated with the hashes of the swapped subtrees\footnote{The incremental update will be faster when the cumulated length of the swapped subtrees is smaller than the length of $T_1$.}. For example, the hash of the child individual from~\autoref{fig:gp-crossover} is obtained by:
\[
    \mathcal{H}(C_1) = \mathcal{H}(T_1) \oplus {\color{green!60!black}\underbrace{h_{-,4} \oplus h_{x_3,5} \oplus h_{x_4, 6}}_{\text{XOR ``out'' $(x_3 - x_4)$}}} \oplus {\color{blue} \underbrace{h_{\centerdot^2,4} \oplus  h_{x_3, 5}}_{\text{XOR ``in'' $x_3^2$}}}
\]

\subsubsection*{Implementation}

We base our implementation on the open-source library Operon\cite{burlacu2020} that contains a high-performance, concurrent implementation of SR in the \CPP{} programming language. The library makes use of a fine-grained concurrency model where recombination and subsequent fitness evaluation take place in parallel, such that new individuals are generated independently in separate logical threads. In order to not diminish the concurrency and scalability of the library, interposing a fitness cache between the recombination and evaluation steps requires that the cache object is able to operate in a synchronized manner -- that is, the underlying hash map object must support concurrent read/write operations.
In order to satisfy there requirements we employ a parallel hash map library\cite{popovitch2019} with the following characteristics:
\begin{itemize}
    \item \emph{closed hashing}: the values are stored directly in a memory array (avoiding memory indirection)
    \item \emph{data-parallel lookup}: using vector instructions, the hash table is able to look up items by checking 16 slots in parallel
    \item \emph{segmented design}: the hash table is internally composed of an array of submaps with internal locking (one mutex is created for each internal submap)
    \item \emph{fine-grained locking}: separate locks for each internal submap provide a speed benefit when compared to single threaded insertion
\end{itemize}

The cache is then implemented as a \texttt{hash\_map} object that maps Zobrist hash values to fitness. Integration with the workflow in~\autoref{fig:gp-workflow} is straightforward -- the logic needs to be slightly adapted to only re-evaluate a solution if it has not been cached before, as seen in~\autoref{alg:cached-fitness}. This is implemented in Operon at the level of the recombination operation. The cache is initially updated with the initial population created by the initialization step.

\begin{algorithm}
    \begin{algorithmic}[1]
        \Procedure{CachedFitness}{$T$}
            \State $h \gets \mathcal{H}(T)$
            \If{$\texttt{cache}(h) = \emptyset$}
                \State $\texttt{cache}(h) \gets \textsc{EvaluateFitness}(T)$
            \EndIf
            \State \Return $\texttt{cache}(h)$
        \EndProcedure
    \end{algorithmic}
    \caption{Cached fitness evaluation}
    \label{alg:cached-fitness}
\end{algorithm}

In addition to the hash map object, the  following components are necessary:

\paragraph{Zobrist table} The table is initialized with random values corresponding to each symbol $\times$ position combination, as illustrated in the right-hand side of figure~\autoref{fig:zobrist-tree-example}.
Its shape will be an $n\times l$ matrix where the number of symbols $n$ and the number of indices $l$ are given by:
\begin{align}
    n &= \underbrace{|\text{binary operators}| + |\text{unary operators|}}_\text{function nodes} + \underbrace{|\text{nullary operators}|}_\text{terminal nodes}\\
    l &= \text{the maximum tree length}
\end{align}
For example, if the primitive set contains $20$ symbols and the maximum tree length is $50$, then the Zobrist table will contain $1000$ values. To avoid hash collisions, we use 64-bit values generated with a high-quality pseudo-random number generator\footnote{We use the excellent general-purpose \texttt{xoshiro256**} generator from \url{https://prng.di.unimi.it}}\cite{blackman2021}.

\paragraph{Zobrist hash function} The hash function iterates over tree nodes in preorder and aggregates the corresponding values from the Zobrist table using the XOR operation:
\begin{align}
    \mathcal{H}(T) = \bigoplus_{i=1}^{|T|} h\left( s_i, i \right)
\end{align}
where $T$ is a tree, $s_i$ is the symbol at position $i$ and $h(s_i, i)$ is a hash retrieval function:
\begin{align}
    h(s_i, i) = \begin{cases}
        \texttt{zobrist}(s_i, i) \oplus \texttt{id}(s_i) & \text{if $s_i$ is a variable} \\
        \texttt{zobrist}(s_i, i) & \text{otherwise}
    \end{cases}
\end{align}
In the case of variables, each Zobrist table value is XOR'd with a unique identifier associated with each variable. This definition of $h$ is necessary in order to treat variables as distinct symbols within this convention such that, for example, $\mathcal{H}({x_1}) \neq \mathcal{H}\left({x_2}\right)$.

\paragraph{Hashing of coefficients}
The Operon framework as well as many other SR implementations assign a multiplicative coefficient to each tree node in order to enable a more granular fine-tuning of the model using local search. Although these node coefficients can be included in the hash by XOR-ing their respective values into the hash, we believe this to be counter-productive as it would greatly decrease the efficiency of the cache. Therefore, we restrict the tree hash to nodes without coefficients, in effect only hashing tree structure, in order to lower memory requirements, prevent the cache from growing very large and avoid storing differently parameterized versions of the same tree structure.

\section{Experimental Results}\label{sec:results}

The purpose of our experiment is to ascertain the effectiveness of fitness caching in reducing runtime while maintaining solution quality. We therefore compare otherwise identical algorithmic configurations with and without fitness caching, using a number of real-world datasets illustrated in~\autoref{tab:test-problems}.
We employ the Non-dominated Sorting Genetic Algorithm 2 (NSGA2)~\cite{deb2002}, a multi-objective evolutionary algorithm whose selection mechanism is based on the concepts of Pareto dominance and crowding distance, to allow the simultaneous optimization of model accuracy and size.
The algorithm is configured with the hyperparameters described in~\autoref{tab:nsga2-settings}. A single best model is selected from the returned Pareto front using the Minimum Description Length principle, which favors models that provide the shortest description of the data in an information-theoretical sense\cite{grunwald2019}, therefore choosing the best compromise between error and complexity.

Since the individuals are cached based only on their structure without accounting for coefficients, we employ variable amounts of local search in order to ensure that model coefficients are properly optimized before fitness is evaluated.
Recent research has shown that due to the Baldwin effect, a relatively small amount of local search is sufficient in GP~\cite{burlacu2024}. Operon performs a non-linear least squares optimization of model coefficients using the Levenberg-Marquardt algorithm\cite{gavin2019}, a Newton step-based gradient descent method where the step is adjusted to remain within a trust region using a damping factor $\lambda$. When the step is successful, the trust region is expanded by reducing $\lambda$. If the step is unsuccessful, the trust region is shrunk by increasing $\lambda$.

It is expected that a sufficient amount of local search will ensure that the fitness values stored in the cache are the best that can be attained for a given individual, countering the effect of hashing only the tree structure. The experiment includes a configuration with no local search (iterations set to zero), in order to analyze the effects of caching in the extreme case where no care is taken about the coefficients of the models.
We perform 100 repetitions for each problem and set of hyperparameters, for a total of 9000 algorithmic runs.

\begin{table}
    \begin{tblr}{
        width=1.0\textwidth,
        colspec={X[l]X[r]X[r]r},
        row{1} = {font=\bfseries}
    }
        Name & Features & Training size & Test size\\
        \hline
        Chemical-I\cite{kordon2008} & 25 & 3136 & 1863\\
        Chemical-II\cite{kordon2008} & 57 & 711 & 355\\
        Flow stress\cite{kabliman2021} & 2 & 4200 & 3600\\
        FrictionDyn\cite{kronberger2018} & 17 & 1010 & 1016\\
        FrictionStat\cite{kronberger2018} & 16 & 1010 & 1016\\
        Nasa Battery 1\_10\cite{saha2007} & 6 & 504 & 132\\
        Nasa Battery 2\_20\cite{saha2007} & 5 & 1101 & 537\\
        Nikuradse-I\cite{nikuradse1950} & 2 & 230 & 132\\
        Nikuradse-II\cite{nikuradse1950} & 2 & 200 & 162\\
    \end{tblr}
    \caption{Test problems}\label{tab:test-problems}
\end{table}

\begin{table}
    \centering
    \begin{tblr}{
        colspec={lX[l]},
        row{1,12} = {font=\bfseries}
    }
    Fixed hyperparameters & Value\\
    Population size & 1000\\
    Function set & $+, -, \times, \div, \exp, \mathrm{logabs}, \sin, \mathrm{sqrtabs}, \mathrm{square}$\\
    Terminal set & $\mathrm{constants}, \mathrm{variables}$\\
    Tree initialization & balanced tree creator\cite{burlacu2020}, max initial tree length = 10 nodes\\
    Crossover probability & 100\%\\
    Mutation probability & 25\%\\
    Mutation operations & insert/remove/replace subtree,\newline change function/variable/coefficient\\
    Maximum generations & 300\\
    Maximum tree length & 20\\
    Maximum tree depth & 10\\
    Variable hyperparameters & Value \\
    Local search iterations & $\{ 0, 1, 10, 50, 100\}$\\
    Use transposition cache & $\{ \text{True, False} \}$\\
    \end{tblr}
    \caption{NSGA-2 Parameters}\label{tab:nsga2-settings}
\end{table}

\begin{table}
    \SetTblrInner{colsep=5pt}
\begin{tblr}{colspec={X[l]ccccc},
        row{1,2} = {font=\bfseries},
        cell{1}{2} = {c=5}{c},
        cell{1}{1} = {c=1}{l},
        row{even}={bg=abstractcolor},
        row{1}={bg=abstractcolor},
        cell{4}{3,4,5,6} = {bg=green!20!abstractcolor}, 
        cell{6}{3} = {bg=green!20!abstractcolor},
        cell{8}{4,5,6} = {bg=green!20!abstractcolor},
        cell{8}{2,3} = {bg=red!15!abstractcolor},
        cell{10,12}{2} = {bg=red!15!abstractcolor},
        cell{10}{3,4,5,6} = {bg=green!20!abstractcolor},
        cell{14}{3,4,5,6} = {bg=green!20!abstractcolor},
        cell{16}{5} = {bg=red!15!abstractcolor},
        cell{16}{2} = {bg=green!20!abstractcolor},
        cell{18}{5} = {bg=green!20!abstractcolor},
        cell{20}{3} = {bg=green!20!abstractcolor}
    }
        Problem            & Local Search Iterations & & & & \\
                           & 0                               & 1                               & 10                              & 50                              & 100 \\
        Chemical-I & $\num{0.8173} \pm \num{0.0826}$ & $\num{0.9098} \pm \num{0.0091}$ & $\num{0.9196} \pm \num{0.0046}$ & $\num{0.9192} \pm \num{0.0046}$ & $\num{0.9199} \pm \num{0.0064}$\\
 & $\num{0.8352} \pm \num{0.0660}$ & $\num{0.9125} \pm \num{0.0062}$ & $\num{0.9220} \pm \num{0.0031}$ & $\num{0.9226} \pm \num{0.0032}$ & $\num{0.9217} \pm \num{0.0035}$\\
        Chemical-II & $\num{0.5527} \pm \num{0.1190}$ & $\num{0.8081} \pm \num{0.0822}$ & $\num{0.8020} \pm \num{0.1207}$ & $\num{0.7881} \pm \num{0.1639}$ & $\num{0.7943} \pm \num{0.1259}$\\
        & $\num{0.5690} \pm \num{0.1279}$ & $\num{0.8270} \pm \num{0.0399}$ & $\num{0.8050} \pm \num{0.0754}$ & $\num{0.8013} \pm \num{0.1290}$ & $\num{0.8066} \pm \num{0.1313}$\\
        Flow-stress & $\num{0.9509} \pm \num{0.0452}$ & $\num{0.9509} \pm \num{0.0426}$ & $\num{0.9986} \pm \num{0.1445}$ & $\num{0.9990} \pm \num{0.0712}$ & $\num{0.9989} \pm \num{0.1174}$\\
        & $\num{0.9435} \pm \num{0.1032}$ & $\num{0.9478} \pm \num{0.0482}$ & $\num{0.9988} \pm \num{0.1010}$ & $\num{0.9990} \pm \num{0.0870}$ & $\num{0.9991} \pm \num{0.0230}$\\
        FrictionDyn & $\num{0.7228} \pm \num{0.0773}$ & $\num{0.8561} \pm \num{0.0254}$ & $\num{0.8648} \pm \num{0.0069}$ & $\num{0.8632} \pm \num{0.0143}$ & $\num{0.8678} \pm \num{0.0069}$\\
        & $\num{0.6349} \pm \num{0.1201}$ & $\num{0.8686} \pm \num{0.0082}$ & $\num{0.8689} \pm \num{0.0061}$ & $\num{0.8722} \pm \num{0.0043}$ & $\num{0.8712} \pm \num{0.0049}$\\
        FrictionStat & $\num{0.2941} \pm \num{0.1862}$ & $\num{0.8427} \pm \num{0.1491}$ & $\num{0.8533} \pm \num{0.0457}$ & $\num{0.8625} \pm \num{0.0164}$ & $\num{0.8584} \pm \num{0.0274}$\\
        & $\num{0.2602} \pm \num{0.2223}$ & $\num{0.8432} \pm \num{0.1513}$ & $\num{0.8524} \pm \num{0.0682}$ & $\num{0.8629} \pm \num{0.0134}$ & $\num{0.8605} \pm \num{0.0215}$\\
        Nasa Bat 1\_10 & $\num{0.9832} \pm \num{0.0086}$ & $\num{0.9922} \pm \num{0.0036}$ & $\num{0.9929} \pm \num{0.0033}$ & $\num{0.9932} \pm \num{0.0039}$ & $\num{0.9928} \pm \num{0.0030}$\\
        & $\num{0.9856} \pm \num{0.0122}$ & $\num{0.9935} \pm \num{0.0015}$ & $\num{0.9938} \pm \num{0.0023}$ & $\num{0.9938} \pm \num{0.0018}$ & $\num{0.9939} \pm \num{0.0018}$\\
        Nasa Bat 2\_20 & $\num{0.9783} \pm \num{0.0678}$ & $\num{0.9724} \pm \num{0.0161}$ & $\num{0.9712} \pm \num{0.0047}$ & $\num{0.9723} \pm \num{0.0042}$ & $\num{0.9716} \pm \num{0.0049}$\\
        & $\num{0.9784} \pm \num{0.0528}$ & $\num{0.9727} \pm \num{0.0040}$ & $\num{0.9710} \pm \num{0.0051}$ & $\num{0.9709} \pm \num{0.0046}$ & $\num{0.9732} \pm \num{0.0047}$\\
        Nikuradse-I & $\num{0.8908} \pm \num{0.1314}$ & $\num{0.9207} \pm \num{0.1888}$ & $\num{0.9863} \pm \num{0.2081}$ & $\num{0.9893} \pm \num{0.1399}$ & $\num{0.9923} \pm \num{0.1253}$\\
        & $\num{0.8817} \pm \num{0.1342}$ & $\num{0.9440} \pm \num{0.1862}$ & $\num{0.9867} \pm \num{0.1326}$ & $\num{0.9925} \pm \num{0.0304}$ & $\num{0.9930} \pm \num{0.0582}$\\
        Nikuradse-II & $\num{0.9728} \pm \num{0.1236}$ & $\num{0.9823} \pm \num{0.0339}$ & $\num{0.9824} \pm \num{0.0199}$ & $\num{0.9824} \pm \num{0.1357}$ & $\num{0.9824} \pm \num{0.0438}$\\
        & $\num{0.9722} \pm \num{0.1564}$ & $\num{0.9824} \pm \num{0.0076}$ & $\num{0.9825} \pm \num{0.0002}$ & $\num{0.9824} \pm \num{0.1171}$ & $\num{0.9824} \pm \num{0.1357}$\\
    \end{tblr}
    \caption{Test performance given as median test $R^2 \pm \sigma$, without caching (above) and with caching (below). Statistically significant differences, computed using a directional Mann-Whitney U test ($\alpha=0.05$) are highlighted with green (better) and red (worse).}\label{tab:results-mdl}
\end{table}

\begin{figure}
    \includegraphics[width=\textwidth]{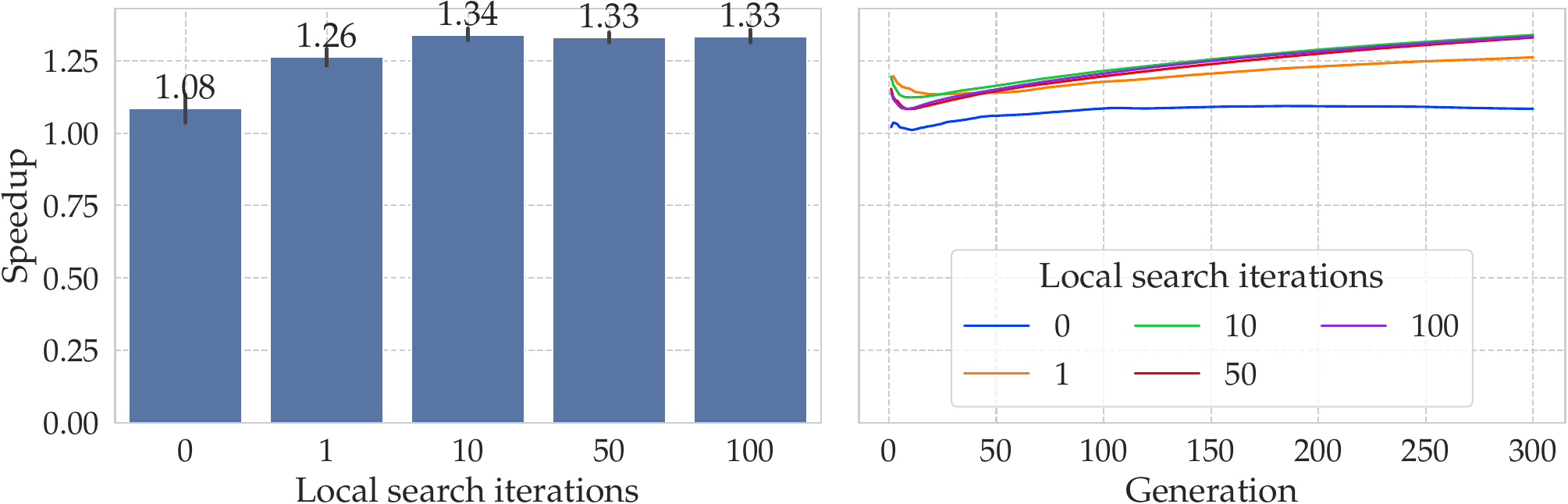}
    \caption{Average speedup obtained with caching at the end of run (left) and generational (right)}\label{fig:average-speedup}
\end{figure}

\begin{table}
    \centering
    \begin{tblr}{colspec={lrrrrrrr},
    row{1,2}= {font=\bfseries},
    cell{1}{4} = {c=5}{c}
}
    Problem & Rows & Columns & Local Search Iterations & & & & \\
    & & & 0 & 1 & 10 & 50 & 100\\
    Chemical-I            & $3136$ & $25$ & $1.17$ & $1.25$ & $1.17$ & $1.21$ & $1.23$\\
    Chemical-II           & $ 711$ & $57$ & $1.09$ & $1.32$ & $1.40$ & $1.31$ & $1.26$\\
    Flow-stress           & $4200$ & $ 2$ & $1.35$ & $1.46$ & $1.28$ & $1.41$ & $1.70$\\
    FrictionDyn-OneHot    & $1010$ & $17$ & $1.10$ & $1.22$ & $1.26$ & $1.22$ & $1.15$\\
    FrictionStat-OneHot   & $1010$ & $16$ & $0.93$ & $1.28$ & $1.37$ & $1.26$ & $1.21$\\
    Nasa Battery 1h 10min & $ 504$ & $ 6$ & $0.95$ & $1.25$ & $1.35$ & $1.29$ & $1.28$\\
    Nasa Battery 2h 20min & $1101$ & $ 5$ & $0.96$ & $1.26$ & $1.41$ & $1.36$ & $1.34$\\
    Nikuradse-I           & $ 230$ & $ 2$ & $1.03$ & $1.13$ & $1.40$ & $1.42$ & $1.37$\\
    Nikuradse-II          & $ 200$ & $ 2$ & $1.17$ & $1.18$ & $1.42$ & $1.49$ & $1.46$
\end{tblr}
    \caption{Average speedup per problem}
    \label{tab:average-speedup-problem}
\end{table}

\begin{figure}
    \begin{subfigure}[b]{0.49\textwidth}
        \includegraphics[width=\textwidth]{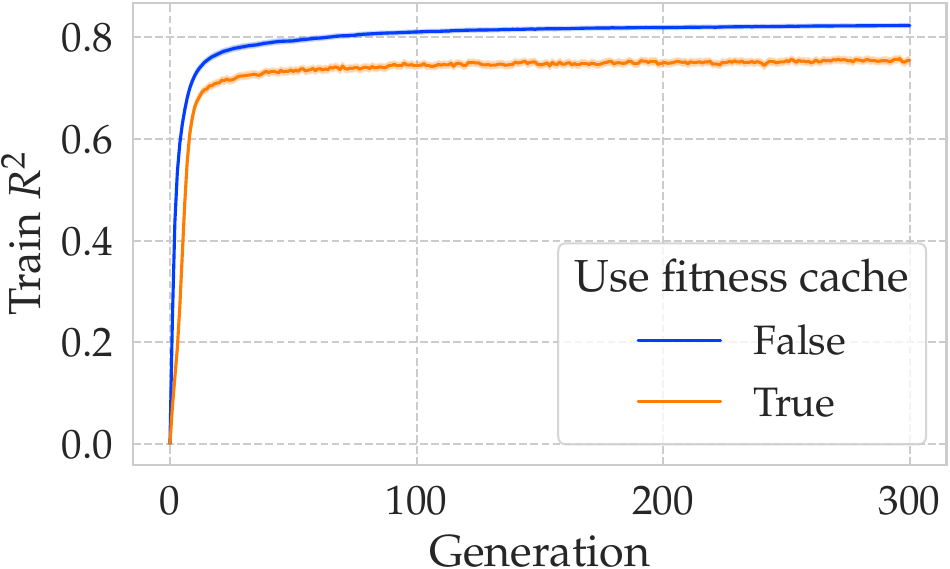}
        \caption{Average fitness}\label{fig:avg-fitness}
    \end{subfigure}
    \begin{subfigure}[b]{0.49\textwidth}
        \includegraphics[width=\textwidth]{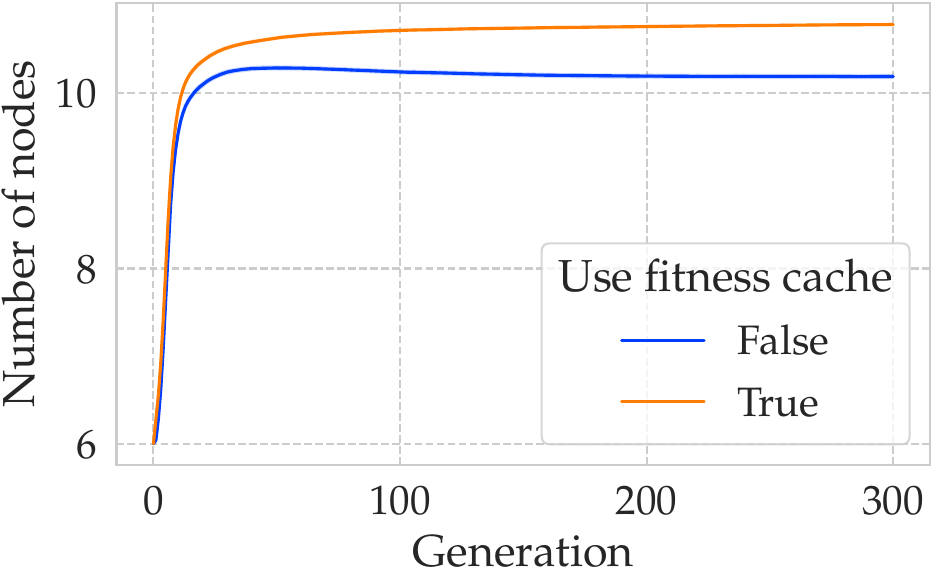}
        \caption{Average length}\label{fig:avg-length}
    \end{subfigure}
    \caption{Average fitness and length, with/without caching. The shaded regions represent the 95\% confidence intervals.}\label{fig:avg-fitness-length}
\end{figure}

\begin{figure}
    \begin{subfigure}[b]{0.49\textwidth}
        \includegraphics[width=\textwidth]{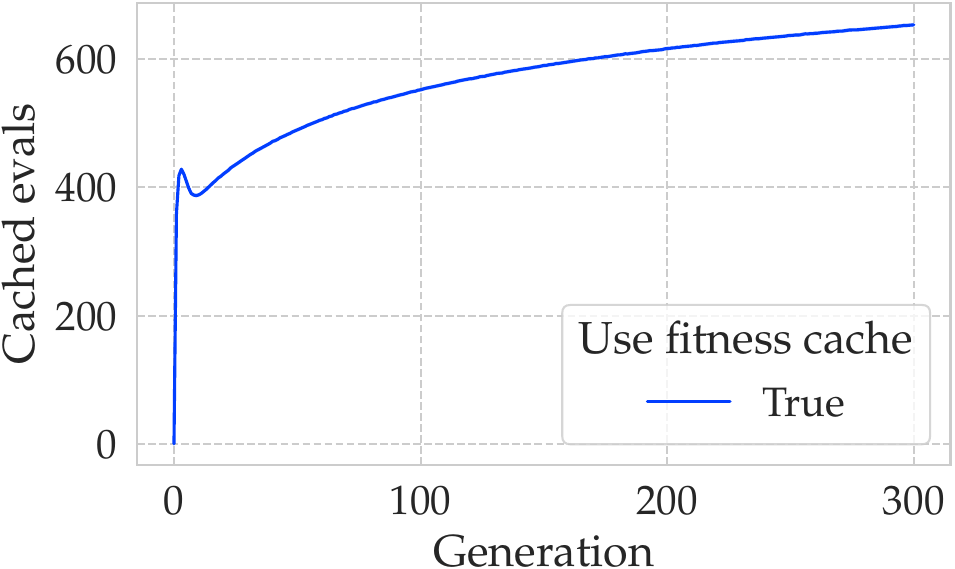}
        \caption{Cached evaluations per generation}\label{fig:cached-evals}
    \end{subfigure}
    \begin{subfigure}[b]{0.49\textwidth}
        \includegraphics[width=\textwidth]{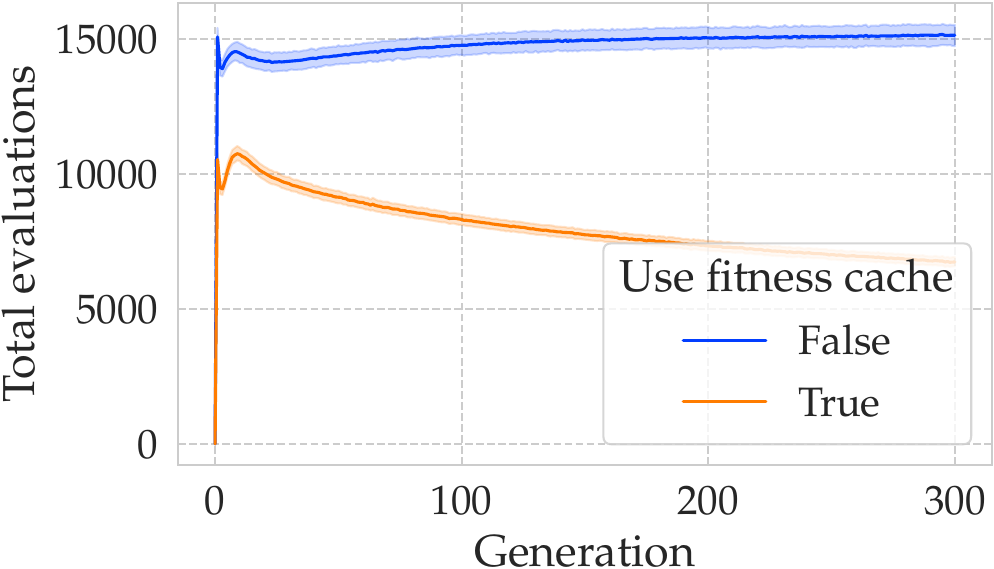}
        \caption{Total evaluations per generation}\label{fig:total-evals}
    \end{subfigure}
    \caption{Cached and total evaluations per generation. The shaded regions represent the 95\% confidence intervals.}\label{fig:evals}
\end{figure}

\paragraph{Results overview}

Experimental results demonstrate the runtime benefits of caching (illustrated in~\autoref{fig:average-speedup}) with observed speedups of up to 34\%. Looking at problem dimensions, we observe a positive correlation ($r=0.42$) between speedup and number of training rows and a negative correlation ($r=-0.44$) between speedup and number of features. Speedup values per problem and number of iterations are illustrated in~\autoref{tab:average-speedup-problem}. The largest measured speedup is 70\% for the Flow Stress problem with $100$ iterations.

At the same time, the median test $R^2$ values (illustrated in~\autoref{tab:results-mdl}) show that caching does not have a detrimental effect on fitness, but rather tends to produce slightly better results. A Mann-Whitney U statistical test performed using the 100 $R^2$ test values obtained for each configuration confirms that, in a majority of cases, no statistically significant difference in the $R^2$ test values could be detected. The SR variant with caching achieves statistically significant better results (albeit with negligible differences in absolute terms) in 19 out of 45 test configurations (highlighted with green in~\autoref{tab:results-mdl}), and statistically-worse results in 5 out of 45 test configurations (highlighted with red). It is worth mentioning that most statistically-worse results occur with an insufficient amount of local search (0 or 1 iterations). A local search amount of 10 iterations per model appears to be optimal in terms of both runtime and model quality. This can be explained by the fact that due to the selection mechanism, the effects of local search in an evolutionary algorithm is cumulative, as good coefficient values are preserved.
However, despite similar results, a more detailed analysis of the microevolutionary dynamics at play reveals a notable difference in search behavior with fitness caching. The full set of results is available online: \url{https://github.com/foolnotion/symreg-london-2025}.

\begin{table}
    \centering
    \begin{tblr}{colspec={rrrr}, row{1,1}= {font=\bfseries},}
    Iterations & Evaluations (Cache=True) & Evaluations (Cache=False) & Saved effort (\%)\\
      0 & \num[scientific-notation=true,round-precision=3]{ 121572.43} & \num[scientific-notation=true,round-precision=3]{ 301000.00} & \num{0.60}\\
      1 & \num[scientific-notation=true,round-precision=3]{ 703541.25} & \num[scientific-notation=true,round-precision=3]{1655112.85} & \num{0.57}\\
     10 & \num[scientific-notation=true,round-precision=3]{2039971.73} & \num[scientific-notation=true,round-precision=3]{4032718.69} & \num{0.49}\\
     50 & \num[scientific-notation=true,round-precision=3]{4093903.56} & \num[scientific-notation=true,round-precision=3]{7275709.48} & \num{0.44}\\
    100 & \num[scientific-notation=true,round-precision=3]{5061250.77} & \num[scientific-notation=true,round-precision=3]{8965539.58} & \num{0.44}
\end{tblr}
    \caption{Impact of local search on total model evaluations computed at the end of the run. Saved effort is computed relative to the not cached quantity.}\label{tab:local-search-impact}
\end{table}

\paragraph{Evolutionary dynamics}

Although the final models are just as good, SR with caching displays slightly higher best model size (overall average 20.46 nodes compared to 20.18 without caching). However, these results are significant only in 8 out of 45 configurations (3 lower, 5 higher best model length with caching).
At the same time, \autoref{fig:avg-fitness-length} clearly shows lower average fitness as well as higher average expression size with caching. A possible explanation for these differences is given by the fact that caching leads to a more uniform fitness landscape which in turn diminishes the effects of selection pressure. Since the NSGA2 algorithm selects for high fitness and low length, the results are lower average fitness and higher average length.
The effects of local search are significant even with a single iteration of gradient, further supporting the hypothesis of the Baldwin effect in GP.

\paragraph{Effects of caching}

We first measure ``cache hits'' (i.e. the number of times an individual's fitness is retrieved from the cache instead of being computed anew) and relate this quantity with the algorithm's population size of 1000 individuals. This indirectly represents a measure of population diversity, since a high cache hit rate implies duplicate individuals. The results shown in~\autoref{fig:cached-evals} show that major diversity loss occurs in the first few generations of the search, followed by a more gradual increase of duplicate individuals. As the amount of duplicate individuals increases, more evaluation effort is saved due to caching, as illustrated in~\autoref{fig:total-evals}. We note that even with a population size of 1000 individuals, a much higher amount of evaluations is possible due to local search. Here, the total includes the computation of residuals and gradients for local search. Even though a large amount of re-evaluations is avoided due to caching, the observed speedup is markedly lower due to other algorithmic steps other than evaluation taking some part of the total runtime, as differences in search dynamics (for example, a higher average length with caching) and overheads of the caching mechanism. The impact of local search on the number of model evaluations (including automatic differentiation and the computation of residuals and Jacobians) is shown in~\autoref{tab:local-search-impact}. We note that the Levenberg-Marquardt algorithm may stop after fewer iterations in case of convergence, which explains the small differences in evaluation effort between 50 and 100 local search iterations. This further supports the idea that a lower amount of local search is sufficient in an evolutionary algorithm.

\section{Conclusion and Future Work}\label{sec:conclusion}

In this work we have shown that significant speedups of the evolutionary search in SR can be achieved via fitness caching. These improvements are made possible through the application of the Zobrist hash, a concept well-known in search algorithms for abstract board games. By indexing node positions and symbol types, the same concept is applicable for creating unique keys for the symbolic expressions that are encountered during the evolutionary search. However, its inability to detect semantic duplicates is an important limitation of this approach, that could be overcome in the future by introducing simplification heuristics for expressions, based on known identities and mathematical equivalences.

Our implementation based on the Operon framework utilizes a highly-optimized concurrent hash map library and is able to achieve up to 34\% speedups (expected to increase with larger datasets) while maintaining solution quality. The high retrieval rate of fitness values from cache also suggests that diversity is rapidly lost in the initial phase of the evolutionary search. We also observe that fitness caching changes search behavior by altering the distribution of fitness values in the population, with notable effects in terms of average fitness and model length. This phenomenon is worth investigating further, as it could also lead to new mechanisms for directing the search and improving overall convergence.

Future research efforts will be focused in finding new applications for the Zobrist hash in other areas of the evolutionary algorithm, notably selection and recombination, such as: dynamic control of selection pressure, selecting for uniqueness alongside fitness and/or complexity, preventing duplicates during recombination or extending the caching concept to sub-expressions.

\bibliographystyle{RS}
\bibliography{references}
\end{document}